\documentclass[a4paper,twoside]{article}
\pdfoutput=1 
\usepackage{epsfig}
\usepackage{subcaption}
\usepackage{calc}
\usepackage{amstext}
\usepackage{amsthm}
\usepackage{multicol}
\usepackage{pslatex}
\usepackage{apalike}
\usepackage{bbm}
\usepackage{multirow}
\usepackage{array}
\usepackage{caption,tabularx,booktabs}
\usepackage{subcaption}
\usepackage{url}     %
\usepackage{paralist} %
\usepackage{SCITEPRESS}     %
\usepackage{makecell}
\usepackage{xcolor}

\begin{document}

\title{
Computer-aided abnormality detection in chest radiographs in a clinical setting via domain-adaptation
\\
\thanks{Biomedical Science, Engineering, and Computing Group, Oak Ridge National Laboratory, Oak Ridge, USA}
\thanks{This manuscript has been authored by UT-Battelle, LLC under Contract No. DE-AC05-00OR22725 with the U.S. Department of Energy. The United States Government retains and the publisher, by accepting the article for publication, acknowledges that the United States Government retains a non-exclusive, paid-up, irrevocable, world-wide license to publish or reproduce the published form of the manuscript, or allow others to do so, for United States Government purposes. The Department of Energy will provide public access to these results of federally sponsored research in accordance with the DOE Public Access Plan (http://energy.gov/downloads/doe-public-access-plan).}
}

\author{\authorname{Abhishek K Dubey\sup{1}\orcidAuthor{0000-0001-8052-7416}, 
Michael T Young \sup{1},
Christopher Stanley \sup{1},
Dalton Lunga \sup{1}, and
Jacob Hinkle \sup{1}}
\affiliation{\sup{1}Oak Ridge National Laboratory, Oak Ridge, TN, USA}
\email{\{dubeyak, youngmt1, stanleycb, lungadd, hinklejd\}@ornl.gov}
}


\abstract{
Deep learning (DL) models are being deployed at medical centers to aid radiologists for diagnosis
of lung conditions from chest radiographs.
Such models are often trained on a large volume of publicly available labeled radiographs.
These pre-trained DL models' ability to generalize in clinical settings is poor because of 
the changes in data distributions between publicly available and privately held radiographs.
In chest radiographs, the heterogeneity in distributions arises
from the diverse conditions in X-ray equipment and their configurations
used for generating the images.
In the machine learning community, the challenges posed by the heterogeneity in the data generation source 
is known as domain shift, which is a mode shift in the generative model.
In this work, we introduce a domain-shift detection and removal method to overcome this problem.
Our experimental results show the proposed method's effectiveness in deploying a pre-trained DL model for abnormality detection in chest radiographs in a clinical setting.
}

\keywords{Computer-aided diagnosis of lung conditions, Domain-shift detection and removal, Chest radiographs}

\onecolumn \maketitle \normalsize \setcounter{footnote}{0} \vfill

\section{Introduction}
\label{sec:intro}
Chest radiography is one of the most ubiquitous diagnostic modalities 
for cardiothoracic and pulmonary abnormalities in the clinical setting.
A timely diagnostic based on the radiographs is a critical step in the 
clinical workflow.
However, many healthcare centers often suffer either from a heavy workload or shortage  
of experienced radiologists.
Deployment of a reliable abnormality detection system would be advantageous in both
scenarios.
Deep learning (DL) based abnormality detection systems are an emerging technology, which is 
yet to be successfully deployed in clinical settings.
The domain shift encountered in privately held datasets due to heterogeneity in 
data generation sources continue to be a prime impediment when deploying pre-trained DL models.
In this work, we introduce a domain-shift detection and removal method to deploy pre-trained DL models in clinical settings.

Domain-shift in this context
is formally defined as the changes in the marginal probability density $p(x)$
between privately held chest radiographs and publicly available radiographs.
The goal of domain-shift detection is to quantify the changes in the
marginal $p(x)$.
While training a model on a public labeled data source $\{x_i, y_i\}_{i=0}^{n}$, 
the best hope is to learn the conditional probability
$p(y|x)$ that is stable or varies smoothly with the marginal $p(x)$.
Even if the conditional is stable, learned models may suffer from model
misspecification, i.e., the learned model may not
perfectly capture the functional relationship between $x$ and $y$ and the approximate solution may become
sensitive to changes in $p(x)$.
The goal of domain-shift removal is to find a transformation $B$ of the data
that minimizes the difference between marginal distributions
of the transformed samples $B(x)$ of privately-held data and public data
to reduce the effect of sensitivity on prediction.
\begin{figure*}
\centering
\includegraphics[width=0.9\textwidth]{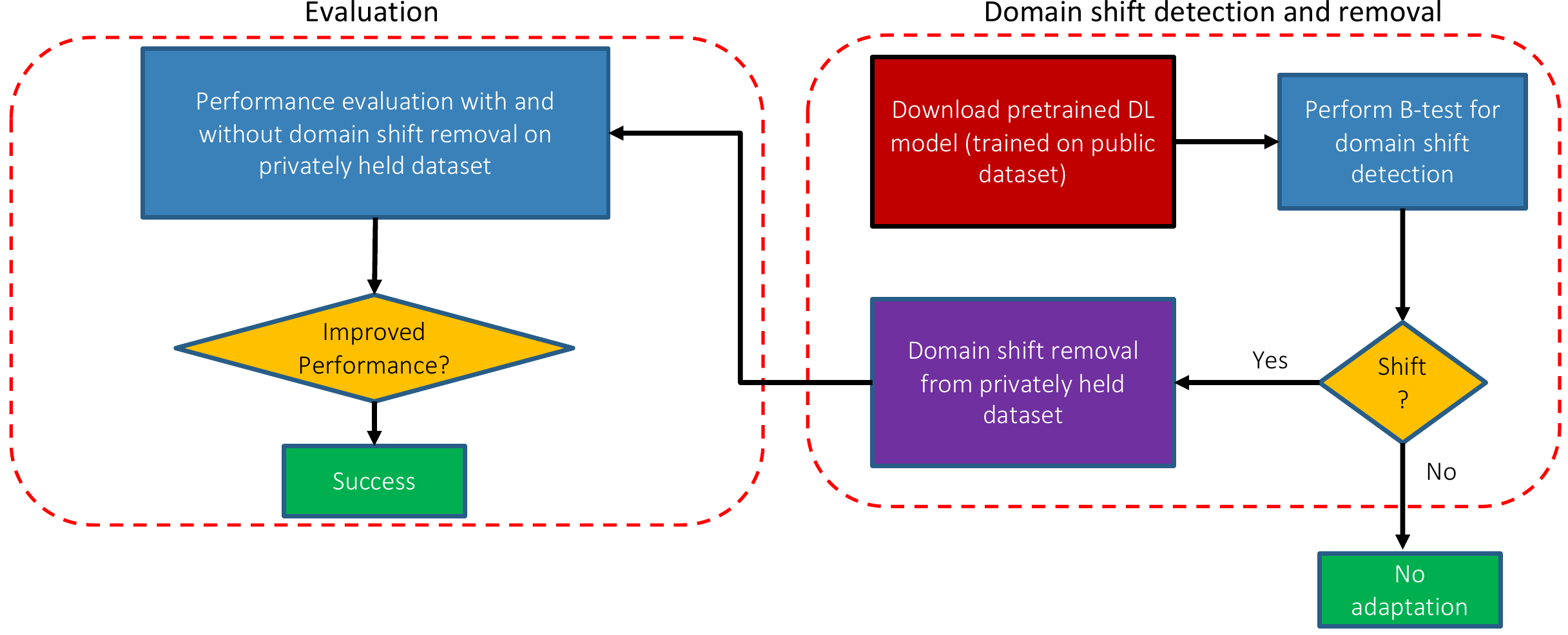}
\caption{Clinical workflow for computer-aided abnormality detection in chest radiographs}
\label{fig:clinical-workflow}
\end{figure*}

Domain separation~\cite{bousmalis2016domain} provides a competing pioneering technique to handle domain-shift. 
Domain separation aims to separate the feature representation
between publicly available and privately held radiographs into domain-invariant and
domain-specific features. 
Domain-shift removal and predictive
modeling are tightly coupled in the domain separation
technique, which requires non-trivial changes
in predictive models to overcome domain-shift.
This paper introduces a novel workflow for deploying the state-of-the-art 
pre-trained DL model for abnormality detection in clinical settings without requiring any changes in
network architecture to overcome domain-shift.
Figure~\ref{fig:clinical-workflow} shows the proposed workflow. 
%
%
In this workflow, we first characterize domain-shift between samples of privately held and public chest radiographs.
In particular, we show that the two sources differ in the distribution of high-frequency components 
such as noise and texture, which we characterize by the density of wavelet scattering transform of radiographs.
Then we learn a generative adversarial network to map samples of a privately held dataset to match their style distribution to that of public chest radiographs.
To evaluate the workflow, we assess the pre-trained DL model's performance on privately held radiographs for abnormality detection 
with and without the domain-shift removal step.

\section{Related work}
\label{sec:related-work}
In this context, one approach is to learn a transformation that embeds data into domain invariant feature space,
which has domain generalization ability to previously unseen domains.
Domain invariant component analysis (DICA)~\cite{muandet2013domain} is among such methods in the 
literature.
%
DICA assumes that data samples come from various unknown distributions and 
it estimates the distributional variance from the data sources.
DICA then finds the orthogonal transform $B$ onto a low-dimensional 
subspace that minimizes the distributional variance while 
preserving the functional relationship between samples and  class-labels.
However, such methods require data samples coming from various unknown distributions
to estimate the distributional variance.

Another method in this category includes domain invariant variational autoencoder (DIVA)~\cite{ilse2019diva}. 
DIVA extends the variational autoencoder framework by disentangling
latent representations for a domain label ($z_{d}$), a class label ($z_{y}$) 
and any residual variations in the inputs ($z_{r})$.
This work claimed to learn a domain-invariant representation using
semi-supervised training utilizing the labeled and unlabeled data from 
both domains.
They used three separate encoders $q_{\phi_{z_d}}(z_d | x)$, $q_{\phi_{z_y}}(z_y| x)$
and $q(\phi_{z_r})(z_r | x)$ and an additional parameterized neural network
$p_{\theta(x|z_d, z_y, z_r)}$ as a decoder.
Their work looks promising for the domain-adaptation task in general.
However, their network architecture has non-trivial differences from the
existing state-of-the-art architecture developed for the abnormality detection
in chest radiographs.

In this work, we propose a workflow to facilitate the use of state-of-the-art 
DL architecture via domain-shift removal from the privately held radiographs.
The domain-shift removal problem broadly falls in the computer vision community 
under the unpaired image-to-image translation category. 
We identify the changes in noise and texture characteristics of radiographs as the main 
difference between the data sources.
We use \textsf{CycleGAN}~\cite{zhu2017unpaired} for removing these differences through
image-to-image translation. 
{\textsf{CycleGAN}} improves upon generative adversarial networks by
exploiting cycle consistency property~\cite{dubey2018iterative,iliopoulos2019idvf,dubey2018symmetric} in the
forward and backward translation maps to avoid mode collapse in the process of image-to-image translation.
%
%

%

\section{Method}
\label{sec:mothods}
\subsection{Domain-shift detection}
\label{ubsec:domain-shift-detection}

\begin{figure*}
\centering
\includegraphics[width=\textwidth]{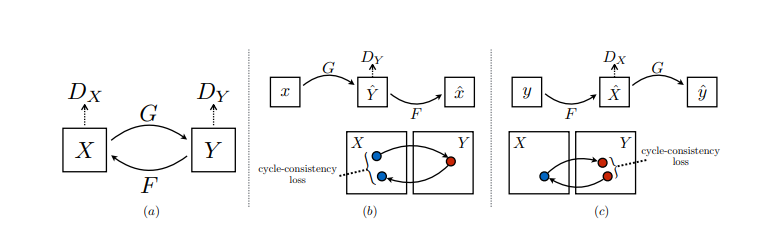}
\caption{CycleGAN contains mappings between two domains $G : X \rightarrow Y$ and $F : Y \rightarrow X$, and
 one discriminator for each domain, $D_X$ and $D_Y$.
The purpose of including discriminators is to encourage the generators $G$ and $F$ to generate samples
that can not be indistinguished with the available real samples from the two domains.
Additionally, CycleGAN introduced cycle consistency losses to enforce forward and backward cyclic consistency between the generators,
i.e., $x \rightarrow G(x) \rightarrow F(G(x)) \approx x$, and
$y \rightarrow F(y) \rightarrow G(F(y)) \approx y$.
We have included this figure into this manuscript from the CycleGAN paper~\cite{zhu2017unpaired}.}
\label{fig:cyclegan-arch}
\end{figure*}

The goal of domain-shift detection is to identify the
shift in the marginal probability density $p(x)$
between two domains $X$ and $Y$ given training 
samples $\{x_i\}_{i=1}^{N}$ and $\{y_i\}_{i=1}^{N}$, 
where $x_i \in X$ and $y_i \in Y$.
We denote the true marginal probability density of
two datasets as $x \sim p_x(x)$ and $y \sim p_y(y)$.  
We formulate the domain-shift detection as 
a hypothesis testing problem, whether
to accept the null hypothesis that there is no domain-shift 
$\mathcal{H}_0: p_{x} = p_{y}$ or to accept the alternative hypothesis 
that there is a domain-shift $\mathcal{H}_1: p_x \neq p_{y}$.
The hypothesis testing often suffers from the curse of dimensionality in 
high-dimensional data settings in estimating test statistic.
In this work, we use a kernel two-sample test initially proposed by~\cite{gretton2012kernel},
which addressed the problem posed by high-dimensional data settings by introducing
the maximum mean discrepancy (MMD) as test statistic.
The MMD is a distance-measure between probability densities and is defined 
as the largest difference in expectations between the two probability distributions
over functions in the unit ball of a suitable reproducing kernel Hilbert space (RKHS).
The MMD can be empirically estimated between the probability density $p_x$ and $p_y$ by 
the squared distance between their mean embeddings in the RKHS as
\begin{eqnarray}
\eta_k(p_x, p_y) & = \|\mu_k(p_x) - \mu_k(p_y)\|_{{\mathcal{H}_k}}^{2}, 
\end{eqnarray}
where $\mu_k(p_x)$  and $\mu_k(p_y)$ are mean
embedding of $p_x$ and $p_y$, and $\mathcal{H}_k$ is an RKHS with reproducing kernel $k$.
%
%
In this work, we use the B-test statistic as an MMD estimate proposed by~\cite{zaremba2013b}. 
The B-test statistics is an MMD estimate obtained by averaging the $\hat{\eta}_k(i)$, where
each $\hat{\eta}_k(i)$ is the empirical MMD based on a subsample of size $B$.
The asymptotic distribution for $\hat{\eta}^{k}$ under $\mathcal{H}_0$ and $\mathcal{H}_1$ are 
shown to be Gaussian in~\cite{zaremba2013b}.
Following~\cite{zaremba2013b}, we set the subsample size $B$ to $\sqrt{n}$
to obtain a consistent estimator.
A user-defined threshold $\alpha$, which denotes the test level, 
is used to determine whether the test statistic is sufficiently large as to 
accept the alternative hypothesis $\mathcal{H}_1$, that is a shift in the marginal
distributions $p_x$ and $p_y$.

In this work, we use a fixed convolutional neural
network called Wavelet scattering transform in composition with the
radial basis function as the kernel function.
The Wavelet scattering transform is used to extract the features that are invariant to translation and 
Lipschitz stable to deformation.
The higher-order wavelet scattering transform is shown to characterize the noise and 
texture in the signal by~\cite{bruna2013invariant}.
We use the scattering transform to capture this high-frequency component of the radiographs essentially,
which is the characteristic difference between the domains.
Then we use the radial basis kernel to map the scattering coefficient to the kernel space to find the B-test statistics.

Next we identify the out-of-distribution (OOD) samples that require domain-shift removal.
For this purpose, we empirically estimate the density of the samples from the source and target domain in a 
low-dimensional subspace spanned by the principal components of the scattering coefficients.
We identify the samples that are in the non-overlapping region between the source and target domains as potential
candidates for the domain-shift removal.
\subsection{Domain-shift removal}
\label{ubsec:domain-shift-removal}

The goal of the domain-shift removal is to learn a mapping $G: X \rightarrow Y$ from the privately held dataset
domain $X$ to publicly available dataset domain $Y$.   
We use the state-of-the-art method, \textsf{CycleGAN}~\cite{zhu2017unpaired}, to perform this task.
{\textsf{CycleGAN}} additionally learns the reverse mapping $F:Y \rightarrow X$ and two
adversarial discriminators $D_X$ and $D_Y$ in conjunction with $F$ from the unpaired
samples from $X$ and $Y$ as shown in Figure~\ref{fig:cyclegan-arch}. 
{\textsf{CycleGAN}} enforces inverse consistency conditions, $F \circ G = G \circ F = \mathbbm{1}$, between
the two maps, where $\mathbbm{1}$ is an identity map.
Additionally, the discriminator $D_X$ is learned to distinguish between the real images $\{x \in {X}\}$ and 
translated images $\{F(y), y \in {Y}\}$ and similarly $D_Y$ is learned to discriminate
between $\{y \in{Y}\}$ and $\{G(x), x \in{X}\}$. 

\section{Experimental setup}
\label{sec:experimental-setup}
%
%
This section describes two chest radiograph datasets,  presents their noise and texture characterization, 
and describes the experimental set-up for abnormality detection.
%

%
\subsection{Dataset Description}
\label{subsec:datasets}
We present abnormality detection results on {\textsf{MIMIC-CXR}} dataset.
{\textsf{MIMIC-CXR}} is a publicly available chest radiograph in Digital Imaging and Communications 
in Medicine (DICOM) format. The diagnosis labels are derived from the radiology reports associated with these images.
The dataset contains radiographs associated with $227,827$ patients
collected at the Beth Israel Deaconess Medical Center between $2011$ 
and $2016$.
The dataset is de-identified to satisfy the Health Insurance Portability and Accountability Act 
requirements, and protected health information are removed. 
We converted the DICOM file format ($16$-bit depth raw format) to JPEG file  format ($8$-bit depth raw format) 
using the pydicom library and  downsample the radiographs to $256 \times 256$ pixels for further  analysis.  
We normalized the dynamic range of the images to $[0,  255]$  by the following  steps:
\begin{inparaenum}[(i)] 
\item subtracting the image pixel values with the lowest pixel value in the image,
\item dividing the image pixel values by the highest pixel value
and multiplying pixel values by $255$ in the image,
\item truncating and converting the result to an unsigned integer.
\end{inparaenum}
Finally, we stored the radiographs in the compressed JPEG format with a quality value of $95$.
We did not perform any filtering or pre-processing of the images before storing them in JPEG format.
We used a pre-trained {\textsf{DenseNet121}}~\cite{tang2020automated} for the abnormality detection,
which was trained on another publicly available dataset, {\textsf{ChestXray14}}~\cite{wang2017chestx}, 
released by the National Institute of Health. 
This dataset provides $112,120$ radiographs from $30,805$ patients in PNG format 
at $1024 \times 1024$ resolution.
The dataset was rigorously screened to remove all personally identifiable information.
The {\textsf{ChestXray14}} radiographs have the same dynamic range of $[0,  255]$.
We downsample the radiographs to $256 \times 256$ pixels to make it consistent with {\textsf{MIMIC-CXR}}.
We use {\textsf{ChestXray14}} to learn an image-to-image translation model between the samples of {\textsf{MIMIC-CXR}} and {\textsf{ChestXray14}}, 
which we use for removing the domain-shift from out-of-the-distribution samples of {\textsf{MIMIC-CXR}}.
We did not perform any filtering or pre-processing to {\textsf{ChestXray14}} radiographs before using it for 
{\textsf{MIMIC-CXR}} to {\textsf{ChestXray14}} translation.
\subsection{Domain-shift characterization}
\label{subsec:domain-shift-characterization}
We characterized the noise and texture of \textsf{MIMIC-CXR} and \textsf{ChestXray14} by computing the distribution of wavelet scattering transforms of the datasets.
We used  \textsf{Scattering2D} method from \textsf{Kymatio} package~\cite{andreux2020kymatio} for computing 
the scattering transform.
We computed up to the second-order of the scattering coefficients
by setting \texttt{max\_order=2}. We set the filter parameters \texttt{J=4} and 
\texttt{L=8} while maintaining other parameters to default values. 
We summed the scattering coefficients over the image domain 
to obtain a translational invariant feature.
This way, we extracted $417$ coefficients for every image in the two datasets.
We whitened the coefficients and reduced their dimensionality
using the principal component analysis and estimated data distribution 
in reduced space. 
We estimated the distribution by binning the coefficient space $[-4, 4] \times [-4, 4]$ 
into $50 \times 50$ bins and counting the samples in every bin for both datasets.
Figure~\ref{fig:cxr-sc-cyclegan} shows the count of samples of \textsf{MIMIC-CXR} and
\textsf{ChestXray14} in the binned area.
A domain-shift between \textsf{MIMIC-CXR} and \textsf{ChestXray14} is evident from the figure.
\begin{figure}[tb]
\centering
\includegraphics[width=0.4\textwidth]{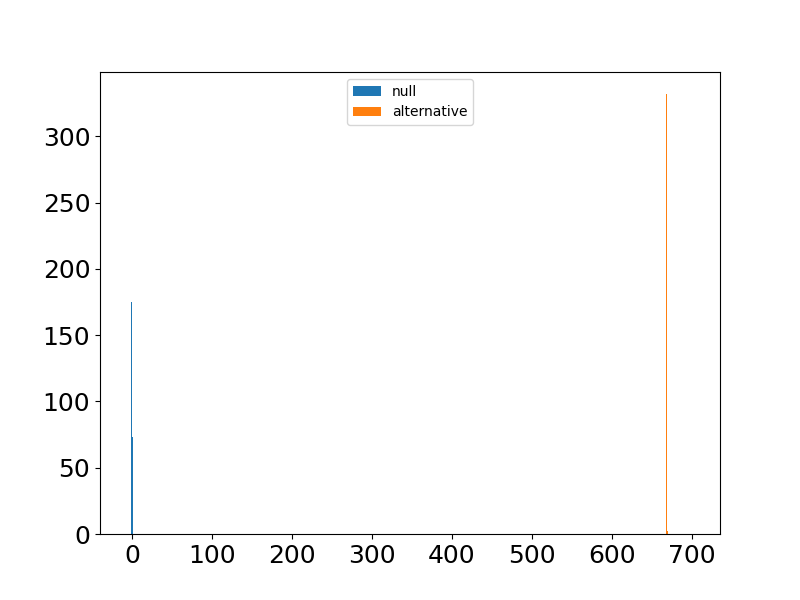}
\caption{ 
Empirical MMD distributions under null ($\mathcal{H}_{0}$) and alternative ($\mathcal{H}_{1}$) hypothesis between 
the ChestXray14 and MIMIC-CXR data sources. 
We used scattering transform in composition with the radial basis function (RBF) as kernel function
in order to find the B-test statistics.
We set \texttt{max\_order=2} to compute up to the second-order of the scattering coefficients 
and set other filter parameters \texttt{J = 4} and \texttt{L = 8}. 
We summed the scattering coefficients over the image domain to obtain a translational invariant $417$ features.
We set the RBF scale parameter \textcolor{gray}{$\gamma=1$}.}
\label{fig:cxr-btest-sc-noadap}
\end{figure}

We performed a two-sample test (B-test)~\cite{zaremba2013b} on the extracted wavelet scattering coefficients with radial basis function kernel.
With the kernel scale parameter \texttt{\textcolor{gray}{$\gamma$}=1}, we get a \texttt{p-value=0} for the two-sample kernel test, indicating an overwhelming support
for the hypothesis that the two datasets come from different distributions.
Figure~\ref{fig:cxr-btest-sc-noadap} shows the supporting statistics in the B-test for the null and alternative hypotheses.

\subsection{Evaluation measures}
\label{subsec:evaluation-measures}
We assess the pre-trained \textsf{DenseNet121}'s performance on the abnormality detection in chest radiographs 
on \textsf{MIMIC-CXR}.
We compare the area under the receiver operating characteristic curve (AUC), 
accuracy, precision, sensitivity, specificity, positive predictive value (PPV), and 
negative predictive value (NPV) scores of the pre-trained model with and without 
domain-shift removal.
We show the class activation maps for some selected examples to aid in the interpretation of \textsf{DenseNet121} results.
We compare the class activation maps of the selected radiographs with and without domain-shift removal to
study the model's sensitivity to noise and texture characteristics in the radiographs.
%
%

\section{Results}
\label{sec:results}
\subsection{Domain adaptation by {\textsf{CycleGAN}}}

\begin{figure}[tb]
\centering
\begin{subfigure}{0.03\textwidth}
\text{\rotatebox{90}{No adaptation}}
\end{subfigure}
\begin{subfigure}{0.21\textwidth}
\centering
\includegraphics[width=0.99\textwidth]{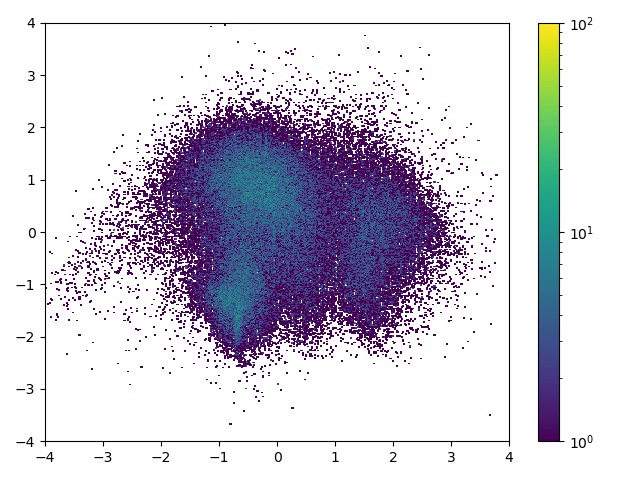}
\end{subfigure}
\begin{subfigure}{0.21\textwidth}
\centering
\includegraphics[width=0.99\textwidth]{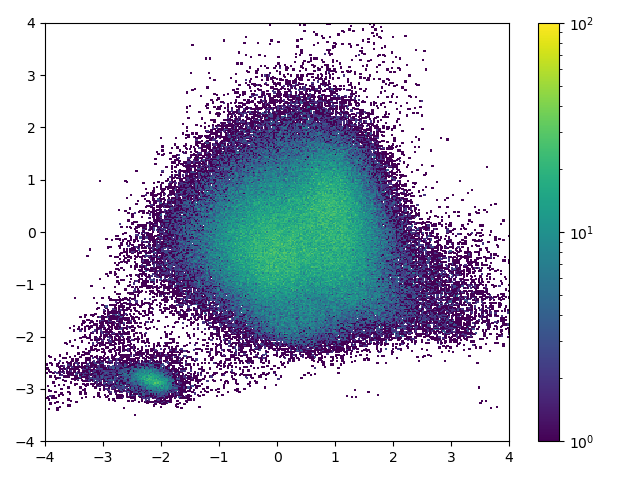}
\end{subfigure}
\\
\begin{subfigure}{0.03\textwidth}
\text{\rotatebox{90}{CycleGAN}}
\end{subfigure}
\begin{subfigure}{0.21\textwidth}
\centering
\includegraphics[width=0.99\textwidth]{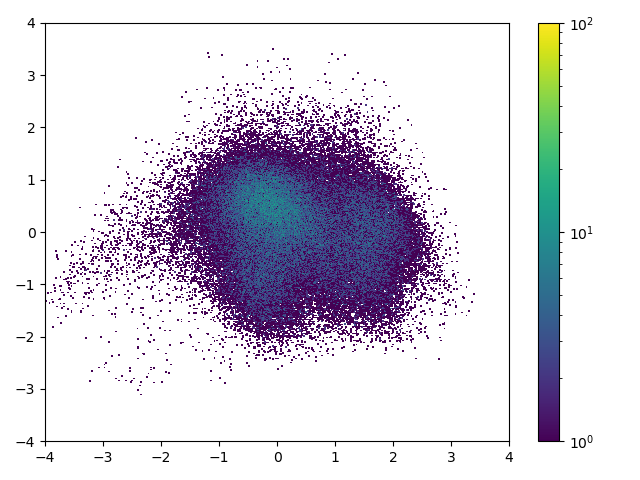}
\caption{ChestXray14}
\end{subfigure}
\begin{subfigure}{0.21\textwidth}
\centering
\includegraphics[width=0.99\textwidth]{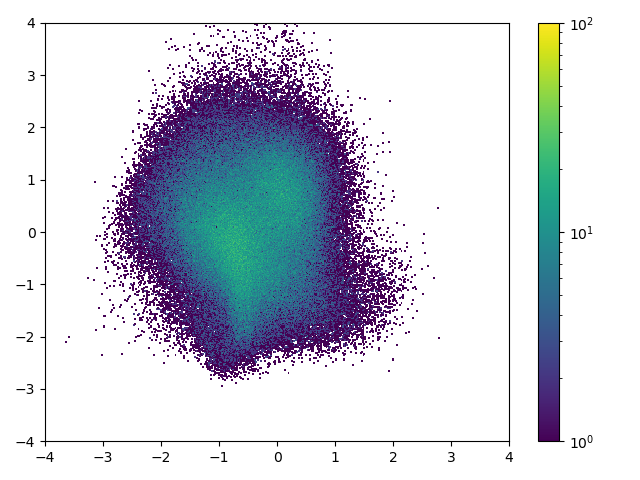}
\caption{MIMIC-CXR}
\end{subfigure}
\caption{
Density plot of two PCA modes of wavelet scattering coefficients (WSCs) of two datasets are displayed.
Top row shows the original distributions of the scattering coefficients,
whereas bottom row shows the distributions after the image-to-image translation
by \textsf{CycleGAN}.
The WSCs are computed using Scattering2D implementation from Kymatio package with parameter setting 
\texttt{J=4}, \texttt{L=8}, \texttt{max\_order=2} while keeping the other parameters to default values.
We summed the scattering coefficients over the image domain to obtain translational invarient features.
We also whitened the features before extracting two PCA components.
}
\label{fig:cxr-sc-cyclegan}
\end{figure}

\begin{figure}[tb]
\centering
\includegraphics[width=0.45\textwidth]{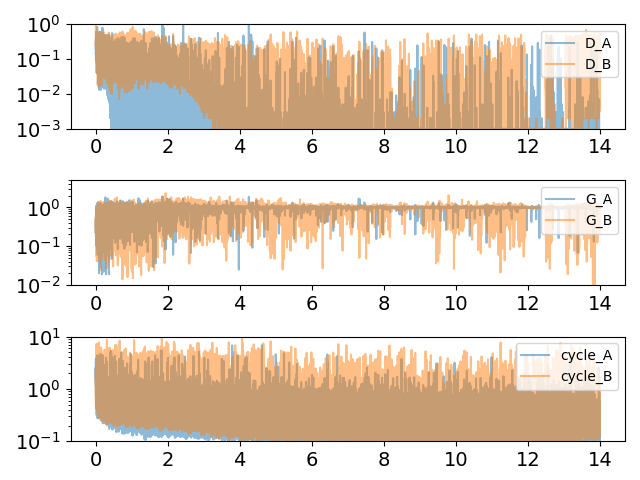}
\caption{
Six training loss of \textsf{CycleGAN}, trained between the samples of \textsf{ChestXray14} and \textsf{MIMIC-CXR}, are displayed for $14$ epochs. 
We denote two discriminator losses by \texttt{D\_A} and \texttt{D\_B}, two generator losses by \texttt{G\_A} and \texttt{G\_B}, and two cycle-consistency losses by \texttt{cycle\_A} and \texttt{cycle\_B}.
}
\label{fig:cxr_cyclegan_training_loss}
\end{figure}

We present the distributions of wavelet scattering coefficients of {\textsf{ChestXray14}} and 
{\textsf{MIMIC-CXR}} dataset before and after the domain-adaptation by {\textsf{CycleGAN}} in 
Figure~\ref{fig:cxr-sc-cyclegan}.
We used the PyTorch implementation\footnote{\url{https://github.com/junyanz/CycleGAN}}
of {\textsf{CycleGAN}}. We used default network architecture and default training and testing 
parameters. Table~\ref{table:cyclegan-param} includes some notable parameters.
We trained {\textsf{CycleGAN}} for $14$ epochs with all $112,120$ {\textsf{ChestXray14}} and selected $243,332$ 
{\textsf{MIMIC-CXR}} radiographs.
The {\textsf{MIMIC-CXR}} radiographs with Posterior-Anterior (PA) 
and Anterior-Posterior (AP) views were selected to keep them consistent with {\textsf{ChestXray14}} source. 
We used the image-to-image translation maps learned by {\textsf{CycleGAN}} to adapt the samples of {\textsf{ChestXray14}} 
and {\textsf{MIMIC-CXR}} to the other domain.
Figure~\ref{fig:cxr_cyclegan_training_loss} reports the training loss of {\textsf{CycleGAN}} 
averaged over $100$ minibatchs with \texttt{batch\_size=1} during entire training process.
Density plots in the figure~\ref{fig:cxr-sc-cyclegan} shows that {\textsf{CycleGAN}} performs well in 
domain adaptation and successfully eliminates most of the non-overlapping areas between the domains.
\begin{table}[tb]
\centering
\caption{
\textsf{CycleGAN} architecture and training parameters.}
\begin{tabular}{p{3cm} p{2cm}}
\toprule
Parameter & Value\\
\midrule
\texttt{netG} & \texttt{resnet-9blocks}\\
\texttt{netD} & \texttt{basic}\\
\texttt{n\_layers\_D} & \texttt{3}\\
\texttt{input\_nc} & \texttt{3}\\
\texttt{output\_nc} & \texttt{3}\\
\texttt{lambda\_A} & \texttt{10}\\
\texttt{lambda\_B} & \texttt{10}\\
\texttt{lambda\_identity} & \texttt{0.5}\\
\texttt{lr} & \texttt{0.0002}\\
\texttt{lr\_policy} & \texttt{linear}\\
\texttt{lr\_decay\_iter} & \texttt{50}\\
\texttt{batch\_size} & \texttt{1}\\
\texttt{no\_dropout} & \texttt{true}\\
\bottomrule
\end{tabular}
\label{table:cyclegan-param}
\end{table}

\begin{table*}[tb]
\caption{
Classification evaluation scores of pre-trained {\textsf{DenseNet121}} on the out-of-distribution (OOD) {\textsf{MIMIC-CXR}} radiographs are compared to the CycleGAN's mapped OOD radiographs' scores in the last two columns. The second column includes {\textsf{DenseNet121}}'s performance on a small {\textsf{ChestXray14}} test cohort reported by a previous study~\cite{tang2020automated}. The third and fourth column contains {\textsf{DenseNet121}}'s evaluation scores on full and $50\%$ abstained MIMIC-CXR datasets. The evaluation scores with abstention are computed on the top $50\%$ of predictions, ranked based on the model's confidence.       
}
\centering
\begin{tabular}{p{2cm} >\centering m{2cm} >\centering m{2cm} >\centering m{2cm} >\centering m{2cm} c}
\toprule
\multirow{3}{*}{Metric}  & ChestXray14 & \multicolumn{4}{c}{MIMIC-CXR}\\
\cmidrule(r){3-6}
& Hold out ($1344$) & Full ($193974$) & Abstention ($96987$) & \multicolumn{2}{c}{OOD (5800)}\\
& & & & No adaptation & Cycle-GAN \\
\midrule
AUC & 0.98 & 0.79 & 0.87 & 0.73 & 0.75\\
Accuracy & 0.95 & 0.79 & 0.90 & 0.71 & 0.74\\
Precision &0.90 & 0.83 & 0.89 & 0.83 & 0.85\\
Sensitivity & 0.97 & 0.82 & 0.96 & 0.60 & 0.63\\
Specificity & 0.93 & 0.76 & 0.79 &0.85 & 0.87\\ 
PPV & 0.90 & 0.83 & 0.89 & 0.83 & 0.85\\
NPV & 0.95 & 0.75 & 0.91 & 0.64 & 0.66\\
\bottomrule
\label{domain-adaptation}
\end{tabular}
\label{table:abnormality-detection-results}
\end{table*}

\subsection{Abnormality detection}

\begin{figure*}[!h]
\centering
\begin{subfigure}{0.47\textwidth}
\includegraphics[width=0.47\textwidth]{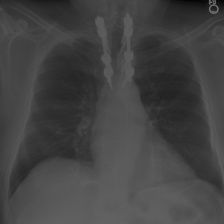}
\includegraphics[width=0.47\textwidth]{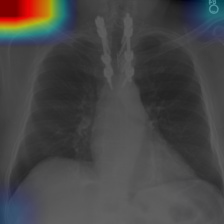}
\end{subfigure}
\begin{subfigure}{0.47\textwidth}
\includegraphics[width=0.47\textwidth]{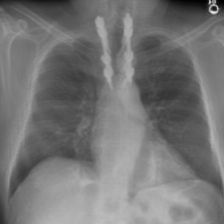}
\includegraphics[width=0.47\textwidth]{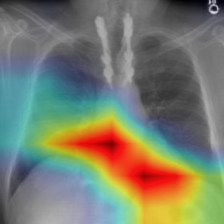}
\end{subfigure}
\begin{subfigure}{0.47\textwidth}
\includegraphics[width=0.47\textwidth]{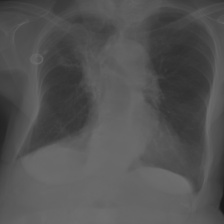}
\includegraphics[width=0.47\textwidth]{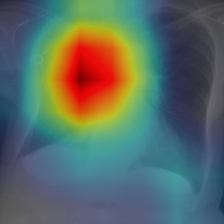}
\end{subfigure}
\begin{subfigure}{0.47\textwidth}
\includegraphics[width=0.47\textwidth]{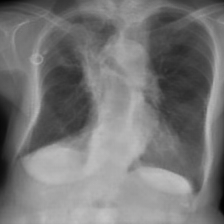}
\includegraphics[width=0.47\textwidth]{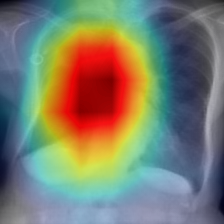}
\end{subfigure}
\begin{subfigure}{0.47\textwidth}
\includegraphics[width=0.47\textwidth]{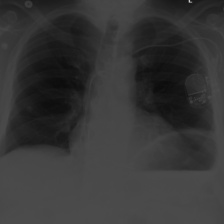}
\includegraphics[width=0.47\textwidth]{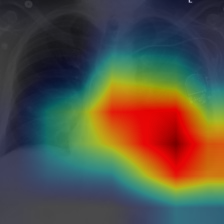}
\caption{Original MIMIC-CXR}
\end{subfigure}
\begin{subfigure}{0.47\textwidth}
\includegraphics[width=0.47\textwidth]{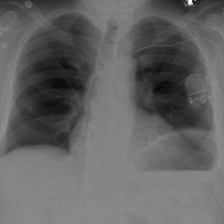}
\includegraphics[width=0.47\textwidth]{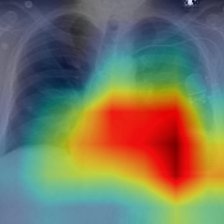}
\caption{Style-adjusted MIMIC-CXR}
\end{subfigure}
\caption{
Left group is the original \textsf{MIMIC-CXR} radiographs and the class activation maps of abnormal regions in the radiograph found by \textsf{Grad-CAM}.
Right group is the translated \textsf{MIMIC-CXR} radiographs obtained by applying \textsf{CycleGAN}'s translation map to the \textsf{MIMIC-CXR} radiographs 
and the class activation maps of abnormal regions in the radiograph found by \textsf{Grad-CAM}. 
}
\label{fig:cxr-cyclegan-images}
\end{figure*}

We present the AUC, accuracy, precision, sensitivity, PPV, and NPV score of the pre-trained \textsf{DenseNet121}~\cite{tang2020automated}
on the abnormality detection task on {\textsf{MIMIC-CXR}} with and without the domain-shift removal.
\textsf{DenseNet121} was trained by the Authors of~\cite{tang2020automated} on \textsf{ChestXray14} dataset using
images with a $256 \times 256$ resolution and was shown to achieve a new state-of-the-art performance
on the abnormality detection binary task.
We downloaded their pre-trained model and tested their model's accuracy on {\textsf{MIMIC-CXR}}  with and without the domain-shift removal.

We derived the labels for the abnormality task detection for the {\textsf{MIMIC-CXR}} dataset.
We set the abnormality label to $1$ when any of the following $6$ conditions are detected 
by both Chexpert~\cite{irvin2019chexpert} and NegBio~\cite{peng2018negbio}:
Cardiomegaly, Consolidation, Edema, Pleural Effusion, Pneumonia, Pneumothorax.
%
We set the abnormality label to $0$ when both Chexpert and NegBio report No Finding.
We exclude all other cases from the {\textsf{MIMIC-CXR}} test cohort. 
This screening process yields a total of $193,974$ labeled radiographs,
including $81,847$ normal radiographs and $112,127$ radiographs with abnormality.

We present our experimental findings in Table~\ref{table:abnormality-detection-results}.
The pre-trained model performs much lower on the full {\textsf{MIMIC-CXR}} than on the small {\textsf{ChestXray14}} test cohort of $1,344$. 
With the abstention of $50\%$, the pre-trained model achieves an accuracy of $90\%$ on {\textsf{MIMIC-CXR}}. 
To calculate the model's performance with abstention, we calculate the model's confidence in abnormality detection 
by calculating $|p\!-0.5|$, where p is the abnormality detection score returned by the pre-trained model.
We ranked the predictions based on the model's confidence, and we kept the top $50\%$ predictions.
Next, we report the pre-trained model's performance on the out-of-distribution (OOD) test sets.
We used the original {\textsf{MIMIC-CXR}} samples that lie in the region $[\!-4,\!-1] \times [\!-4,\!-2]$ in 
Figure~\ref{fig:cxr-sc-cyclegan} as the OOD test set.
We report an improvement of $3\%$ accuracy on the OOD test set with domain-shift removal.  

\subsection{Grad-CAM to study sensitivity to domain-shift}
\begin{figure}[tb]
\centering
\includegraphics[width=0.45\textwidth]{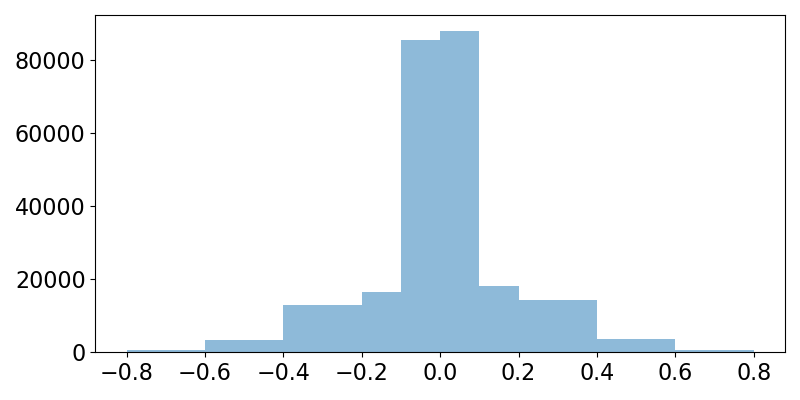}
\caption{
Differences in the abnormality prediction scores obtained by {\textsf{DenseNet121}} on \textsf{MIMIC-CXR} 
and style-adjusted \textsf{MIMIC-CXR} by {\textsf{CycleGAN}} is binned into non-overlapping intervals, 
and the counts in every interval are displayed.
}
\label{fig:cxr_preds_compare_hist}
\end{figure}

We show the class activation maps computed by the {\textsf{Grad-CAM}}~\cite{selvaraju2017grad}
for some selected examples to aid interpretation of {\textsf{DenseNet121}} results.
For each examples, we show the original radiographs of {\textsf{ChestXray14}} and {\textsf{MIMIC-CXR}}, 
adapted radiographs by the {\textsf{CycleGAN}} to the other domain, a heatmap overlaid on the images indicating 
the prediction of abnormal regions by the {\textsf{Grad-CAM}}.
Examples suggest that the {\textsf{DenseNet121}} model is potentially focusing on clinically meaningful 
abnormal regions of the chest radiographs for the classification task, however it is sensitive
to the noise and texture of the input radiographs, as seen in Figure~\ref{fig:cxr_preds_compare_hist}.

\section{Discussion}
\label{sec:discussion}
We hypothesized a distributional difference in noise and texture characteristics between the data sources due to diverse conditions in X-ray equipment and their configurations for generating the images.
We based our hypothesis on the recent findings~\cite{pooch2019can,yao2019strong}, which implicitly showed
the existence of some characteristic differences between these data sources. 
This work has developed an explicit method to show the characteristic differences
between the data sources.
The density-plot of the high-order wavelet scattering transform of these radiographs 
confirms our hypothesis.
We exploited the unpaired image-to-image translation method, CycleGAN, 
to remove this shift and experimentally validated its effectiveness in domain-shift 
removal. 
Our findings also should be applicable to 
discerning unique, private features from common, public ones, which could facilitate 
more targeted privacy-aware DL approaches to best balance privacy-utility.
We also showed that the state-of-the-art model for abnormality detection is susceptible to model 
misspecification and is sensitive to input distribution changes when trained on a single data source.
This finding is consistent with the literature work for other diagnostic tasks~\cite{pooch2019can,yao2019strong}.
We have decoupled the domain-shift removal and model construction 
due to the applicability of such a decoupled method to various downstream tasks.
However, a problem-specific coupled solution to abnormality detection with adversarial training 
is also possible, which is out-of-scope of this paper.
Our main contribution in this work is the introduction of distribution of high-frequency components 
to characterize the data sources and relating it to the difficulty of pre-trained models to generalize
on unseen domains.
In this work, we have introduced a framework for domain-shift detection and removal to overcome this problem.

\section{Acknowledgements}
This research is sponsored in whole or in part by the AI Initiative (LOIS $9613$) and Privacy research (LOIS $9831$) as part of the Laboratory Directed Research and Development Program of Oak Ridge National Laboratory.

\nocite{*}
\bibliographystyle{apalike}
{\small
\bibliography{references}}

\end{document}